# A Language and Its Dimensions: Intrinsic Dimensions of Language Fractal Structures

Vasilii A. Gromov, Nikita S. Borodin, and Asel S. Yerbolova

HSE University, Pokrovskii boulevard, 11, Moscow, 101000, Russia

Correspondence should be addressed to Vasilii A. Gromov; stroller@rambler.ru

## Abstract

The present paper introduces a novel object of study – a language fractal structure; we hypothesize that a set of embeddings of all $n$-grams of a natural language constitutes a representative sample of this fractal set. (We use the term *Hailonakea* to refer to the sum total of all language fractal structures, over all $n$). The paper estimates intrinsic (genuine) dimensions of language fractal structures for the Russian and English languages. To this end, we employ methods based on (1) topological data analysis and (2) a minimum spanning tree of a data graph for a cloud of points considered (Steele theorem). For both languages, for all $n$, the intrinsic dimensions appear to be non-integer values (typical for fractal sets), close to 9 for both of the Russian and English language.

## 1 Introduction

Gromov and Migrina [1] seem to take a fresh look on a natural language to prove that it is a unified complex system, more preciously, it is a self-organized critical system, with texts (literature masterpieces) corresponding to avalanches. The present paper examines geometric properties of the system. To this end, for a given language, we consider the set of all embeddings (for all words [$n$-grams] of the language) as a unified geometric object. We assume that a set of all embeddings (of observed words) of the language constitute a representative sampling of all points of a $d$-dimensional surface. Therefore, one can explore the set of all embeddings in order to ascertain geometric features of this surface. The same holds true for the surfaces associated with the sets of all observed $n$-grams. Looking ahead, we have to say that these surfaces feature non-integer (fractal) dimensions, so we have to abandon the first term that suggests itself, 'language manifolds,' in favour of 'language fractal structures.' The authors propose to use the term *Hailonakea*[1] ('the sign immensity' from Hawaiian) to refer to the totality of all language fractal structures, for all $n$-grams, $n$=1, …, a coarse-graining structure of a language as a complex system [2].

---

[1] A reader may click the link https://en.wikipedia.org/wiki/Laniakea_Supercluster in Wikipedia to see a photography of Laniakea Supercluster, the largest object of the observable Universe. Each pixel of the photograph corresponds even not to a galaxy, but to a cluster of galaxies; the photograph gives an insight into the coarse-graining structure of our Universe. "The name *laniākea* ([ˈlɐnijaːˈkɛjə]) means 'immense heaven' in Hawaiian, from *lani* 'heaven', and *ākea* 'spacious, immeasurable'" (Wikipedia). Alternatively, one may use the term Lanihailona (the sign heavens) or Semiakea (semiotics + Laniakea). The authors are indebted to Nikolai Yu. Yutanov, the Central Astronomical Observatory of the Russian Academy of Sciences at Pulkovo (Russia), and his colleagues, The Mauna Kea Observatories (USA), who help coin the term.

The present paper discusses the intrinsic dimensions[2] of the language fractal structures for $n$=1 (words); $n$=2 (bigrams); $n$=3 (trigrams) for the Russian and English languages. The problem in question seems to be theoretically[3] important: it introduces a new mathematical object – a language fractal structure – and examines its fundamental properties.

To run the simulations in order to estimate the intrinsic dimensions, we employ the corpora of the literatures for the Russian and English language. The reasons why we chose these very languages are: (1) both languages enjoy large corpora of literature texts; (2) they are inherently different, in a sense: the Russian language possesses rich inflections and free word order [5], whereas the English one is an isolating language, and it tends to follow the strict rules on word order [6]. The authors believe, maybe, somewhat quaintly, that the literature constitutes a core of the respective language. Consequently, we think that it is quite reasonable to use $n$-grams extracted from the corpora of literature masterpieces in order to investigate the respective language.

The rest of the paper is organized as follows. The second section reviews the related works, the papers that serve as a starting point for the present one. The third introduces the problem statement. The fourth discusses methods employed to estimate the intrinsic dimensions; the fifth, the estimated intrinsic dimensions for synthetic data (in order to verify methods performance and robustness); the sixth, the estimated intrinsic dimensions for Russian and English fractal language structures. Finally, the last section presents conclusions.

## 2 Related works

It seems to be helpful to approach the available literature from two angles. First, we review papers that examine an entire language as a natural-science object standpoint[4]; then, those that examine an intrinsic dimension of a given set of points. Importantly, this study proceeds from the belief that a natural language constitutes an array of meanings (concepts), rather than that of words (symbols), and one should treat it as a semantic space (space of meanings).

Unfortunately, most papers that study a natural language as a whole concern themselves with 'words', rather than 'meanings'. Nevertheless, we may indicate several monographs [1, 8, 9] that explore 'words' in order to reveal language processes properties for 'meanings'. Tanaka-Ishii and his colleagues ([9] and references therein) analyse long correlations for the English and Japanese languages. Due to the nature of Japanese language, one can establish nearly one-to-one correspondence between hieroglyphs (kanjis) and meanings. Seemingly, one can cautiously extend conclusions made for hieroglyphs to elements of a semantic space ('meanings'). The language demonstrates long correlations, those of words placed 10-15 positions apart[5]. Dębowski [8] investigates the power laws of a natural language: he demonstrates that this very class of distributions such processes governs most fundamental language processes. Gromov and Migrina [1] examines a natural language as a whole unity; the authors prove that a natural language constitutes a self-organised critical system, with a literature masterpiece (or any other text) constituting a power-law governed 'avalanche' in a semantic space. As a rare exception to papers that deal with 'words', we may refer to the paper [10]: the authors transform words into embeddings, thereby transforming literature texts into multivariate chaotic series.

---

[2] We formally define the intrinsic dimension below, in the latter sections. For the time being, to gain an insight on the intrinsic dimension, one may just consider a sphere – its intrinsic dimension is equal to 2 (as one can uniquely locate a point on its surface using two values, its latitude and longitude), regardless the dimension of the space one embeds the sphere (3, 5, etc.). We should stress that, if the one samples points from the sphere surface, one will obtain a set of vectors with dimensions equal to that of the space one embeds the sphere into (3, 5, etc.). Respectively, we will mistake the sets of vectors for a 3-,5-dimensional manifold – meanwhile its true (intrinsic) dimension is equal to 2 in any case.

[3] On the other hand, the study casts some doubts on whether it is worthwhile to utilize embeddings of huge dimensions in NLP, as LLM tend to do, especially those based on deep models (BERT [3], OpenAI GPT [4], and so forth).

[4] Apparently, this goal drives N. Chomsky to establish several variants of generative grammar [7].

For algorithms to estimate an intrinsic dimension, first, we should refer to the paper [11]: the paper introduces necessary conditions for a function to be an intrinsic dimension function; the respective axioms rely on M. Gromov's distance between metric spaces [12]. One can broadly separate the available methods to estimate an intrinsic dimension of a geometrical object (using a representative sample of its points) into three classes. They, in order to estimate an intrinsic dimension, employ, respectively, strange attractors for trajectories flowing over the manifold in question [13, 14]; data (proximity) graphs [15]; persistent homologies [16, 17].

The first-class papers examine semantic and sentiment time series of literature masterpieces (see, for example, [10] and references therein). One assumes that a time series at hand satisfies Takens's theorem; this means that it moves in the neighbourhood of the strange attractor to be explored, allowing one to estimate attractor dimensions. Kantz and Schreiber [13], and Malinetsky and Potapov [14], summarise the main ways of defining the strange attractors dimensions (topological, Hausdorff, entropy ones) and main methods of estimating them. Refs [18, 19] concern themselves with ways to estimate fractal dimensions of real-world complex systems. Unfortunately, these methods (1) require time series much longer than any available 'language' time series [14, 20]; besides that, they are frequently not robust with respect to changed data.

Costa et al. [21] and Farahmand et al. [22] employ a $k$-nearest neighbour graph to estimate a topological dimension. Brito et al. [15] extend the method to other data (proximity) graphs.

Adams et al. [16] employ persistent homology to calculate the intrinsic dimensions; Schweinhart [17] rigorously analysed this estimate, relating it to the entropy dimension. Then Schweinhart extends this method to fractal sets, that is, to sets of non-integer dimensions. A preprint [23] utilizes a Schweinhart-estimated intrinsic dimension of separate texts in order to distinguish texts written by humans and those generated by bots; the authors, as far as we can judge, do not pose problem to analyse a natural language as a whole.

## 3   Problem statement

One can define the dimension of a given set in a variety of ways, stemming from various mathematical disciplines [14]. In what follows, we shall employ the following definitions of the geometrical object:

---

[5] This significantly exceeds the length of typical $n$-gram in natural language processing procedures.

1. Topological (The Lebesgue covering) dimension of the metric space [24] $(X; \delta)$, with a distance function $\delta$, is defined as: $d_T = \inf \{n \in \mathbb{N} | \forall \epsilon > 0 \exists \cup(X; \epsilon, n + 1)\}$; $\cup(X; \epsilon, n + 1)$ is a finite open $(n + 1)$-multiple $\epsilon$-cover of $X$.
2. Hausdorff dimension [25] is defined as: $d_H = \sup \{p : \sup m(\epsilon, p) > 0 | \epsilon > 0\}$; $m(\epsilon, p) = \inf \sum_{\{A_i\} \in \cup(X; \epsilon, n)} (\text{diam } A_i)^p$, $\text{diam } A_i < \epsilon$.

The first definition does not allow a geometric object to possess non-integer values of the dimension[6], whereas the second one does.

The purpose of the present paper is to estimate the intrinsic dimension of language objects. The intrinsic dimension $\partial$, according to Pestov, is defined as [11, 12] the mapping $\partial(X; \delta, \mu) \to \mathbb{R}$ of a space $X$ equipped with a metric $\delta$ and measure $\mu$ (mm-space) that satisfies axioms of concentration, smooth dependence on datasets and normalization[7]. Thus, for each 'language' set under study, we try to estimate its topological and Hausdorff intrinsic dimensions.

Formally, one should:

- For a given set of texts $\mathfrak{I} = (\Omega_1, \cdots, \Omega_N)$ of a given natural language, construct sets $\aleph_n(d)$ of embeddings of all unique $n$–grams, $n = 1..N, d = 1..D$ ($d$ stands for a dimension of embedding space; $n$, for the number of words in the $n$–grams considered). One assumes that the set of texts $\mathfrak{I}$ constitutes a representative sample of texts written in the respective language;
- For a given $n = k$, with the employment of the sets $\aleph_k(d), d = 1..D$, estimate the intrinsic dimensions of $k$–grams $\hat{d} = f_i(\{\aleph_k(d)\})$ of the respective languages, using various methods $f_i$;
- Test the hypothesis that the sets of $n$–grams of a given natural language, $\aleph_n(d)$, are fractal ones, with the employment of the methods $f_i$ able to yield non-integer values of $\hat{d}$.

## 4 Numerical methods to estimate the intrinsic dimension

### 4.1 Geometric representation of a natural language

**SVD.** To construct set of points $\aleph_n(d)$ for a given set of texts (a context) $\mathfrak{I}$, we employ singular value decomposition (SVD) of the co-occurrence matrix [26]:

For a context $\mathfrak{I} = (\Omega_1, \cdots, \Omega_N)$ and a set of words $\aleph = (\lambda_1, \cdots, \lambda_M)$, we construct $M \times N$ matrix $W = (w_{ij})$, with its elements calculated as:

$$w_{i,j} = (1 - \varepsilon_i) \frac{n_{i,j}}{\sum_{i' \in \Omega_j} n_{i',j}}. \tag{4.1.1}$$

$n_{i,j}$ is the number of entries of the (lemmatised) word $\lambda_i$ into the text $\Omega_j$; $\varepsilon_i$ is the normalised entropy of the word $\lambda_i$ in $\aleph$:

---

[6] When applied to a geometric object of non-integer dimension, such algorithms yield the nearest integer value.
[7] Axiom of: (1) concentration: for a family of spaces $(X_n)$ with the form $(X; \delta, \mu)$ $\partial(X_n) \uparrow \infty$ then and only then when $(X_n)$ forms a Lévy family; (2) smooth dependence on datasets: if the M. Gromov's distance $d_{conc}(X_n, X) \to 0$ then $\partial(X_n) \to \partial(X)$; 3) normalization: $\partial(\mathbb{S}^n) = \Theta(n)$, $\mathbb{S}^n$ is an $n$-dimensional sphere.

$$\varepsilon_i = -\frac{1}{\log N} \sum_{j=1}^{N} \frac{n_{i,j}}{\tau_i} \log \frac{n_{i,j}}{\tau_i}, \qquad (4.1.2)$$

$$\tau_i = \sum_{j=1}^{N} n_{i,j}. \qquad (4.1.3)$$

Respectively, the maximum elements of $W$ correspond to the words frequently occurring in the particular text and rarely occurring in all other texts of this context.

For the matrix $W$, thus constructed, we calculate its singular decomposition [27]:

$$W = U\Lambda V^T. \qquad (4.1.4)$$

$U$ and $V^T$ are, respectively, $M \times d$ and $d \times N$ rectangular matrices; $\Lambda$ is a $d \times d$ square diagonal matrix, with singular values occupying its diagonal in the descending order. To calculate SVD, we employ Golub-Kahan-Lanczos algorithm [28]. Rows of the matrix $U$ constitute embeddings of the words ℵ.

Advantageously, in order to obtain $d_2$-dimensional embedding from $d_1$-dimensional one, $d_2 < d_1$, one should truncate it, up to the first $d_2$ components [27]. More to the point, the truncated embedding would retain as much information as possible under such. This fact is extremely important for the present study: to obtain robust results, we should run many simulations, for various embedding dimensions. With this method, it is enough to construct a set of embeddings for large $d$, and then truncate them, if necessary, for smaller $d$-s. Other available methods (for example, deep neural networks) would have required carrying out computationally prohibitive simulations to construct embeddings for each $d$ separately. To obtain embeddings for $n$–grams, we concatenate embeddings of its constituents.

**Word2Vec (CBOW).** Another method we use to obtain embeddings is a Word2Vec-approach Continuous Bag of Words (CBOW) [29].

Vector representations are extracted from the probability distribution resulting from training the a fully-connected neural network. The main idea of CBOW can be described in a few steps: (1) passing through the corpus with a sliding window, (2) maximizing the log-likelihood of the central word depending on the context: the model is a perceptron with one hidden layer, to the input of which word embeddings are transmitted; (3) the output returns a vector of the probability distribution of the belonging of some element hidden in the window to each word in the dictionary. In this way, a word can be predicted from its surrounding context.

Thus, the resulting matrix $W'$ (matrix of weights on the hidden layer) is a dictionary of embeddings (a column vector is assigned to each word). The advantage of Word2Vec is the fact that algebraic operations on vectors reflect semantic operations [29] and cosine proximity of vectors means semantic similarity. At the same time, training the Word2Vec matrix is significantly less time-consuming compared to many other options for constructing language models, and therefore it is chosen as an auxiliary method in this work.

### 4.2  Intrinsic dimension estimation

To estimate the intrinsic dimension of a set of words or $n$–grams for the natural language under investigation[8], we apply two methods. Both methods rely on data graph constructed for the sample considered, and its minimum spanning tree [15, 17].

**Schweinhart estimator.** The first method, discussed in this section, employs Schweinhart's theorem [17], extending Steele theorem [30]:

**Theorem (Schweinhart).** *Let $\mu$ be a d-Ahlfors regular measure on a metric space, and let $\{x_n\}_{n \in \mathbb{N}}$ be i.i.d. samples from $\mu$. Let also the value of the random variable $E_\alpha^0$ be calculated on these samples:*

$$E_\alpha^0(x_1, \ldots, x_n) = \sum_{e \in T_n(\{x_n\}_{n \in \mathbb{N}})} |e|^\alpha. \tag{4.2.1}$$

*Here $|e|, e \in T$ – is a weight value calculated as the Euclidean distance between the vertices of an edge e belonging to the minimum spanning tree $T(x_1, \ldots, x_n)$. Then, if $0 < \alpha < \partial$, so:*

$$C_1 \leq \frac{E_\alpha^0(x_1, \ldots, x_n)}{n^{1-\alpha/\partial}} \leq C_2, \tag{4.2.2}$$

$$\frac{log(E_\alpha^0(x_1, \ldots, x_n))}{log(n)} \to \frac{\partial - \alpha}{\partial} \tag{4.2.3}$$

*for $n \to \infty$; $C_1$ and $C_2$ are positive independent on $n$ constants.*

The theorem makes it possible to estimate (fractal) dimension $\hat{d}_{\text{Schw}}$ of a geometrical object, using a sample of its points $\{x_j\}_{j \in \mathbb{N}}, x_j \in \mathbb{R}^d$ (Fig. 1): one calculates values of $E_\alpha^0$ for various $n$ to form a sample to solve the following regression problem:

$$\ln(E_\alpha^0(x_1, \ldots, x_n)) = \ln(C(\alpha, d)) + \ln(n)\frac{d - \alpha}{d}, \tag{4.2.4}$$

One employs the non-linear MLS method to solve the problem in order to estimate the Hausdorff intrinsic dimension as upper-box dimension ([17], p. 3), using $\frac{d-\alpha}{d} \approx \frac{\hat{d}_{\text{Schw}} - \alpha}{\hat{d}_{\text{Schw}}}$.

**Brito estimator.** The second method [15] we use to estimate the (topological) intrinsic dimension utilises random variables dependent on other kinds of data graphs, those of $k$-nearest neighbours and of influence spheres. Namely, to estimate the intrinsic dimension, we use a sum of squares of vertices degrees for a minimum spanning tree. A large-scale simulation reveals that the statistics is robust[9]: added noise component and (or) increased dimension of embeddings space $d$ do not significantly change the resultant estimates (see Appendix A).

The algorithm is as follows: for a sample of points $\{x_n\}_{n \in \mathbb{N}}$ of a $d$-dimensional space $(\mathcal{X}_d, \delta, \mu)$ equipped with a metric function $\delta$ and a probability measure $\mu$, one:

---

[8] A set of embeddings constructed according to the above procedure.
[9] Brito et al. also proposed two other statistics to estimate the intrinsic dimension in ref. [15]. Unfortunately, these two statistics appeared to be much less robust for the data in hand.

1. Constructs a pairwise Euclidean distance data graph and its minimum spanning tree $T(x_1, \ldots, x_n)$.
2. Calculates a statistic [10]

$$M_n(d) := \frac{1}{n} \sum_{X_i \in T(\{x_n\}_{n \in \mathbb{N}} \subset \mathcal{X}_d)} (\deg(X_i))^2, \quad (4.3.1)$$

3. Estimates parameters [11] of this distribution $f_{prior}(M_n | d^* = i) \approx N(\hat{\mu}_i, \hat{\sigma}_i^2)$ as:

$$\hat{\mu}_i := \frac{1}{L} \sum_{j=0}^{L} M\left(i; \{x_n\}_j \sim U(\mathbf{0}_i, \mathbf{1}_i)\right), \quad (4.3.2)$$

$$\hat{\sigma}_i^2 := \frac{n}{L-1} \sum_{j=0}^{L} \left(M\left(i; \{x_n\}_j \sim U(\mathbf{0}_i, \mathbf{1}_i)\right) - \hat{\mu}_i\right)^2. \quad (4.3.3)$$

$L$ stands for the number of samples, each of $n$ points, drawn from $i$-dimensional unit hypercube.

4. Calculates probability that the desired estimate $\hat{d}$ is equal to $i$

$$p\{\hat{d} = i\} = p(i | M'_n) = \frac{N\left(M'_n; \hat{\mu}_i, \frac{\hat{\sigma}_i^2}{n'}\right)}{\sum_j N\left(M'_n; \hat{\mu}_j, \frac{\hat{\sigma}_j^2}{n'}\right)}, \quad (4.3.4)$$

using Bayes' theorem for $M'_n$.

5. Calculates the estimate of the intrinsic dimension $\hat{d}_{BQY}$ as the mathematical expectation $E[\hat{d}]$, rounded to the nearest integer:

$$\hat{d}_{BQY}(M'_n) = round(E[\hat{d}]) = round\left(\sum_i p(i | M'_n) * i\right). \quad (4.3.5)$$

To control estimates thus obtained, we employ the method based on a formal concept analysis [32, 33]. Its authors admit that, for real-world data, it gives rather rough estimates; nonetheless, one can use it to control an order of the obtained estimates.

## 5 Intrinsic dimensions for conventional manifolds and fractal sets

First, we apply the above algorithms to geometrical objects (manifolds and fractal sets) with a priori known dimensions, embedded into spaces of various embedding dimensions (For list of the objects, refer to Fig. 1.). Each sample consists of 100000 points. In order to increase the dimension of the embedding space [15], we introduce additional coordinates by applying polynomial or periodic function[12] to existing coordinates.

---

[10] Brito et al [15] (refer also to Steele at al. [30]) prove that the average degree of a vertex for the tree remains constant with increase of $n$.

[11] Provided a sequence of random variables $M_n$ converges, with $n \to \infty$, to a random variable with the Gaussian distribution (p. 268).

[12] For each coordinate.

## 5.1 Schweinhart estimator

The algorithm furnishes both point and interval estimates. The latter makes it possible to separate adequate and inadequate estimates: if one considers the dependence of the estimated intrinsic dimension on $\alpha$ (the parameter of the algorithm), one should take into account the point estimates such that their confidence interval does not exceed $\gamma$ of the estimate itself (white area on the respective figures), and discard all others (grey areas on the figures)[13]. Hereinafter, all results are presented for adequate estimates only.

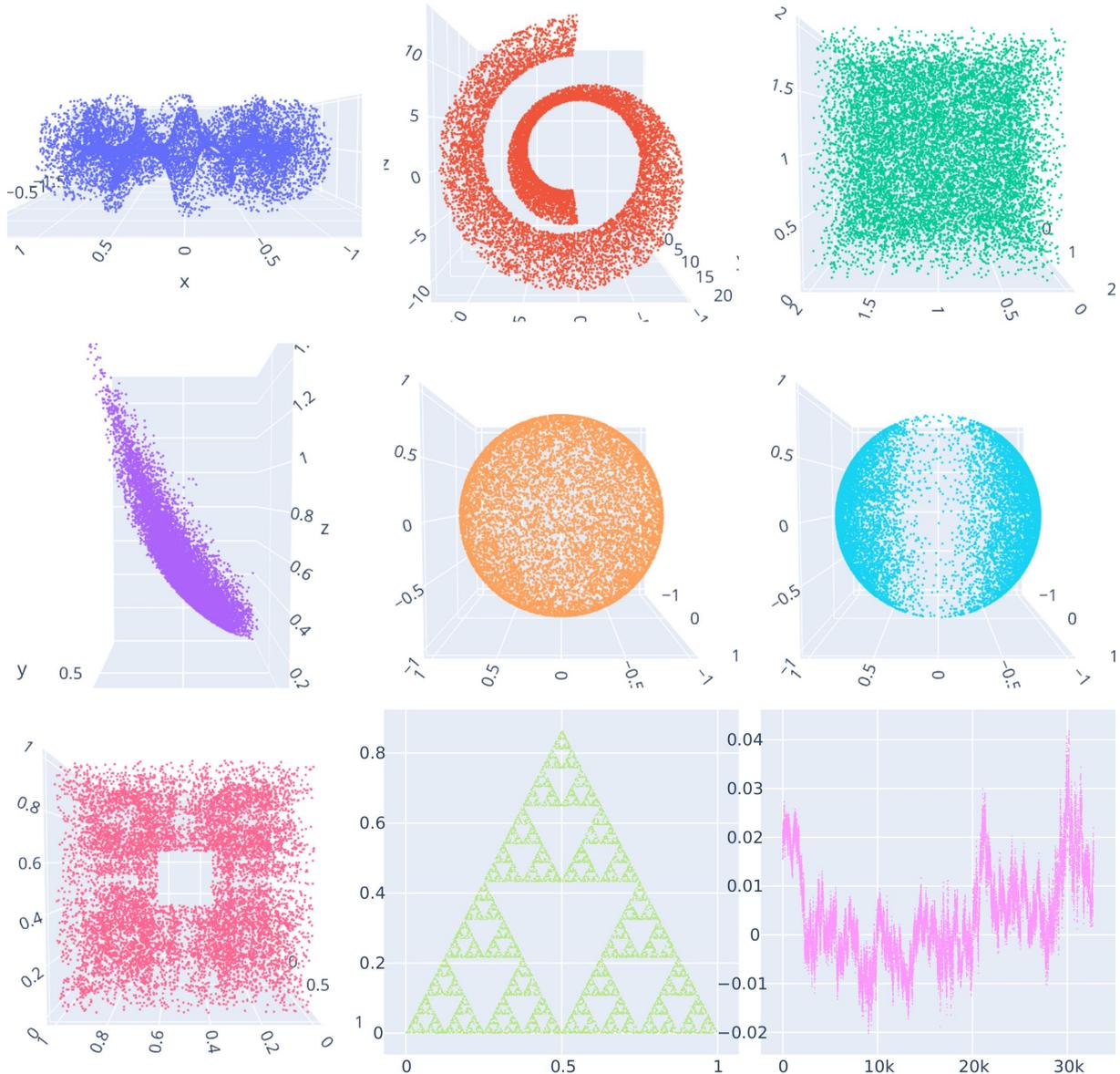

**Fig 1. Conventional geometric objects that we employ to assess the methods of estimating the intrinsic dimension).** The points are distributed uniformly over the surface, unless the otherwise is specified explicitly. Manifolds (the geometrical object of integer Hausdorff dimension): the uppermost row: a Möbius strip, a Swiss roll, a unit cube; the middle row: a paraboloid, a unit sphere, a unit sphere (the points are distributed non-uniformly – a mixture of two Gaussian distribution with modes corresponding to two poles of the sphere). Fractal and multi-fractal sets: the lowermost row: the Menger sponge, the Sierpiński carpet, the log-normal wavelet cascade [34] (a multifractal; time stamps are plotted along the abscissa axis; a state variable, along the ordinate one.

---

[13] $\gamma$ is taken to be 10%.

First, we investigate the dependence of the estimate on $\alpha$. Figure 2 presents typical dependences[14] $\hat{d}_{\text{Schw}}(\alpha)$. Each figure exhibits both a point estimate (a blue solid line) and 95% confidence intervals (red dashed lines). Table A1 in Appendix A summarises the results: the mean average percentage error for the estimated intrinsic dimension does not exceed 5% for all objects. The synthetic data simulation, for objects with known dimensions, reveals that any geometrical object, either regular or fractal, possesses the range of $\alpha$ such that one obtains accurate intrinsic dimension estimates for α from this range. For too small or too large values of $\alpha$, the algorithm yields highly inaccurate estimates of the intrinsic dimension. Fortunately, this range also demonstrates small width of confidence interval for the regression curve. This allows us to formulate several criteria for admissible alpha in order apply them to geometrical objects with a priori unknown intrinsic dimension.

In general, experiments on a significant number of synthetic geometric objects made it possible to identify several characteristic curve types for different types of objects: a horizontal straight line, asymptotically corresponding to an integer value and a non-integer value of the intrinsic dimension for manifolds and fractals, respectively (Fig. 2A and 2B); arcuate curve for multifractals. In all cases, there was a discrepancy between the confidence intervals of the parameter estimate with unlimited growth of $\alpha$; one can attribute it to the fact that the condition    is not satisfied for the respective values [17] (Fig. 2). For multifractals, the dependence 'the estimated intrinsic dimension vs. $\alpha$' (Fig. 2c) exhibits: (1) typical maximum and minimum (or an inflection point); (2) nearly linear dependence. The latter does not reflect the genuine dependence on $\alpha$, but it is an artefact of the method used [16, 17] $\alpha$. Also, the respective range of $\alpha$ exhibits a rapidly increasing confidence interval.

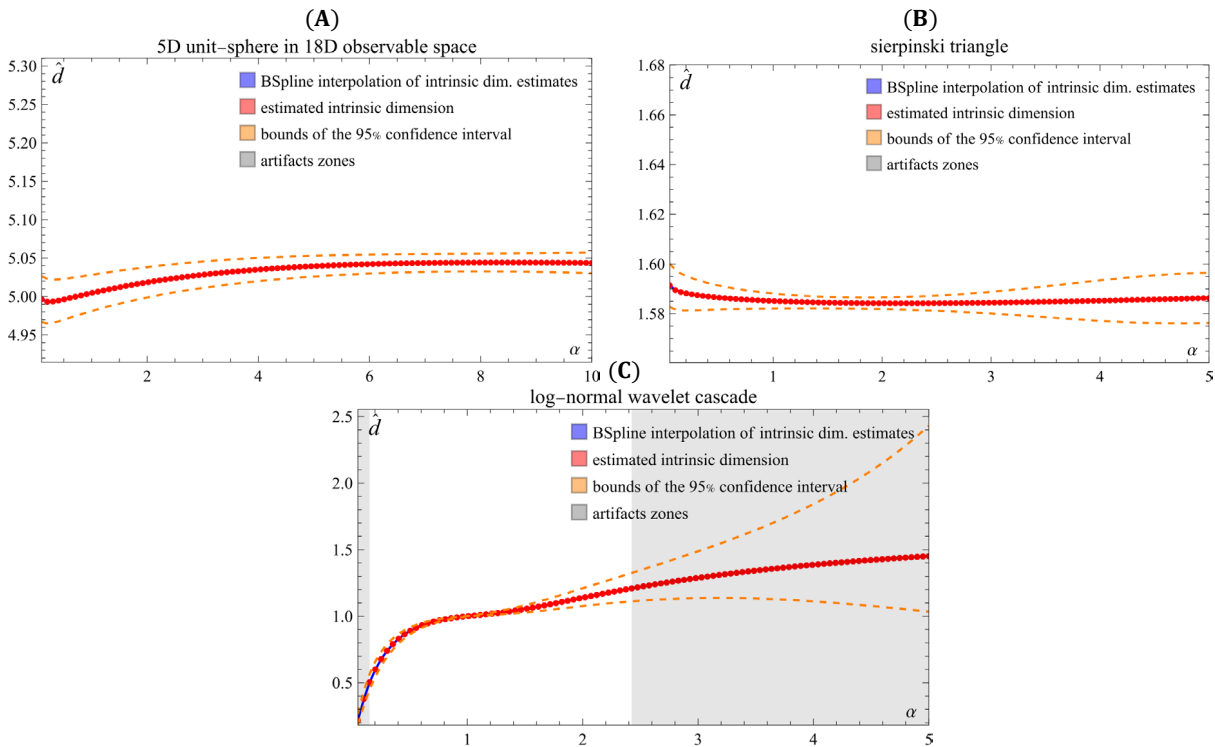

**Fig 2. Dependence of Schweinhart algorithm estimates on parameter α for:** (A) unit sphere (a manifold, $d_H = 5$, $d = 18$); (B) Sierpiński triangle (a fractal set, $d_H = 1.58$, $d = 2$); the log-normal wavelet cascade [34] (a multifractal). The red dots represent the algorithm's estimates, the blue line is a B-spline interpolation of the dependence of the estimates on α, and the orange dotted line represents the upper and lower limits of the 95% confidence interval. The figure also visualizes the criteria for the informativeness of the parameter α, such as the discrepancy of confidence intervals (grey areas).

---

[14] For a simple unit sphere (a manifold, integer dimension), the Sierpiński triangle (a fractal set, non-integer dimension), and lognormal wavelet cascade [34] (a multi-fractal set, non-integer dimension).

The second series of tests was aimed at checking the model's resistance to noise. Noise was introduced in two ways: by adding a random variable with uniform $\sim U(\mathbf{0}, \mathbf{1})$ and normal isotropic $\sim N\big(\mathbf{0}, \mathbf{1} \cdot \mathbf{C}(\text{experiment})\big)$ distributions to each coordinate of the sample point. The results are also shown in the Appendix B. We found the algorithm resistant to noise.

### 5.2 Brito estimator

We conducted similar simulations for the second algorithm. Figure 3 demonstrates a typical scatter plot for this method: the orders of minimal spanning trees are plotted along the abscissa axis, and estimates of the intrinsic topological dimension are plotted along the ordinate axis. Table A2 in Appendix A shows the average results of testing the algorithm on synthetic data (for conventional geometric objects). Evidently, the accuracy of the algorithm decreases with the ratio of the intrinsic topological dimension of the object to that of the embedding space[15]. This means that, for a finite real-world sample, one can use this algorithm just to control results obtained by the first approach.

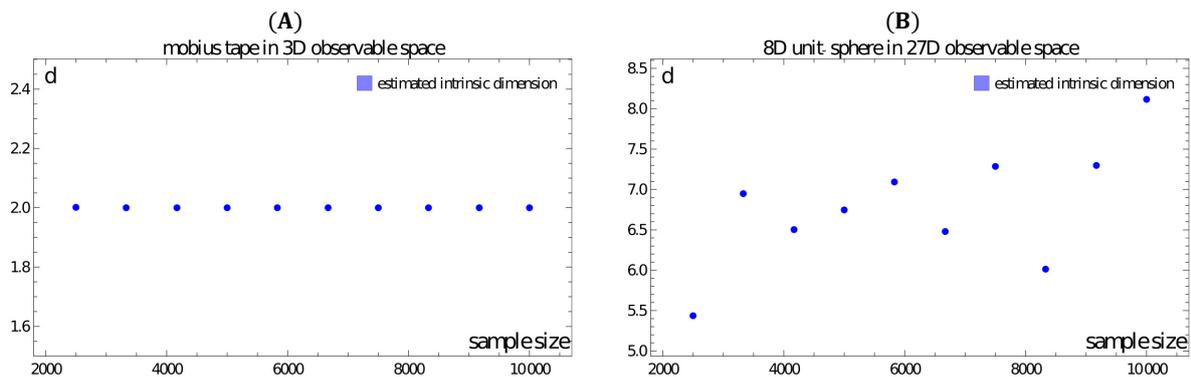

**Fig 3. Dependence of Brito algorithm estimates on sample size for:** (A) mobius tape (a manifold, $d_T = 2$, $d = 3$); (B) unit sphere (a manifold, $d_T = 8$, $d = 27$). As the ratio of the intrinsic dimension to the dimension of the embedding space decreases, all other things being equal, the accuracy of the algorithm decreases too.

## 6 The intrinsic dimensions of the Russian and English languages

To obtain uni- and bigrams embeddings for the Russian and English languages, we employ the corpora of the national literatures, downloadable from open sources. The total number of texts for the Russian language ($|\mathfrak{I}_1|$) amounts to 6429 texts, with 103952 unique words and 14775439 unique bigrams. The total number of texts for the English language ($|\mathfrak{I}_2|$) amounts to 11052 texts, with 94087 unique words and 9490603 unique bigrams.

To preprocess a corpora text, we:
- delete stop-words (articles, conjunctions, expletives, parenthical words, etc.);
- tokenise it (transform proper nouns, numeral, etc. into the appropriate class labels);
- lemmatise words[16] (single out roots);

---
[15] Appendix C shows that large sample size can mitigate this effect: with increasing sample size, the estimates converge to a unified value.
[16] To lemmatise Russian texts, we employ a lemmatise natasha; to lemmatise English ones, a lemmatiser spacy.

To estimate intrinsic dimension of the language fractal structures for embedding dimensions we took $d = \{5, 10, 15\}$. An essential guideline for us was the research on the application of the method of analysing the entropy-complexity pairs [10]: according to the results of the article, for $n = 2$ admissible values for $d$ do not exceed 15.

For the Schweinhart algorithm, for a given language, we
1) generate a sequence of the spanning trees, for $n$ ranging from 1e+5 to the dataset size;
2) for each $n$, for $\alpha$ ranging from 1e-4 to 10 with a step $s \approx 0.1$, we estimate regression (4.2.4) parameters to calculate $\hat{d}_{Schw}$;
3) discard inadmissible $\alpha$ according to the following criteria: (1) the confidence intervals for the regression line are too large (more than $\gamma$); (2) the confidence intervals for parameter $\frac{d-\alpha}{d}$ are too large (more than $\gamma$);
4) select minimal and maximal estimates $\hat{d}_{Schw}$ for all admissible $\alpha$.

For the Brito algorithm, for a given language, we
1) for $d^* \in 2..15$ generate samples $\{x_1^j, ..., x_{1e+6}^j\}_{j=1..100}$, $x_i^j \sim U(\mathbf{0}_{d^*}, \mathbf{1}_{d^*})$, here $\mathbf{0}_{d^*}, \mathbf{1}_{d^*}$ are $d^*$-dimensional zero and unit vectors, respectively;
2) estimate $\hat{\mu}_{d^*}$ and $\hat{\sigma}_{d^*}^2$;
3) calculate estimates $\hat{d}_{BQY}(\{x_1, ..., x_i\}, x_i = [M]_i)$.

Tables 1 and 2 summarise the results. Figures 4 and 5 show typical dependences for estimates employed (see other Figures in Appendices D and E for SVD and CBOW, respectively); they exhibit typical multifractal dependence (please, compare with Fig. 2). The Schweinhart estimator of intrinsic dimensions yields non-integer results, typical for multifractals. The estimated intrinsic dimensions for both languages appear to be quite robust: one gets approximately similar results for various dimensions of embedding spaces[17], of methods to construct embeddings (SVD, CBOW), and methods to estimate intrinsic dimension. This allows one to conclude that intrinsic dimension of a language is a characteristic of a language itself, and, geometrically, the language is multifractal. The fact that the phase space of natural language belongs to the class of multifractals allows us to put forward the following hypotheses:

- language as a dynamic system is a strange attractor [1, 14];
- natural language texts – chaotic series [1, 10].

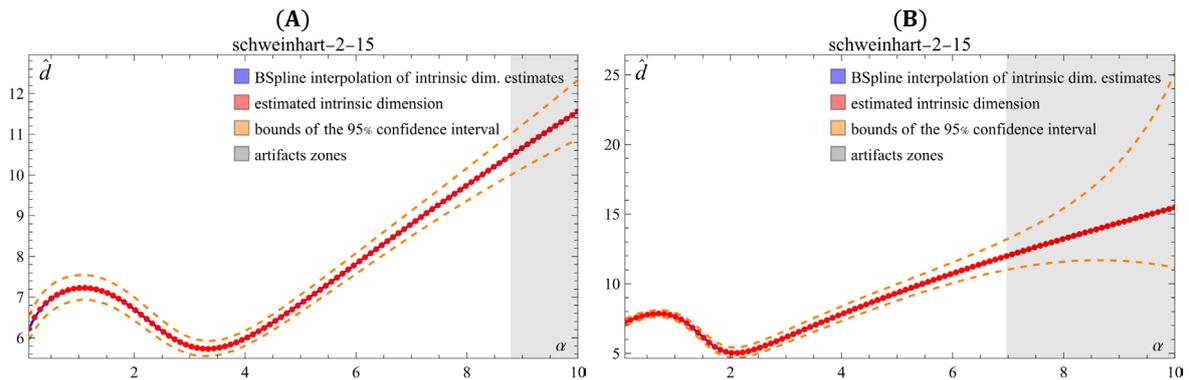

**Fig 4. Dependence of Schweinhart algorithm estimates on parameter α for:** (A) SVD vector representations of Russian national literature ($n = 2, d = 15$); (B) SVD vector representations of English national literature ($n = 2, d = 15$). The red dots represent the algorithm's estimates, the blue line is a B-spline interpolation of the dependence of the estimates on α, and the orange dotted line represents the upper and lower limits of the 95% confidence interval. The figure also visualizes the criteria for the informativeness of the parameter α, such as the discrepancy of confidence intervals (grey areas). The behaviour of graphs for languages is identical to that for multifractal structures.

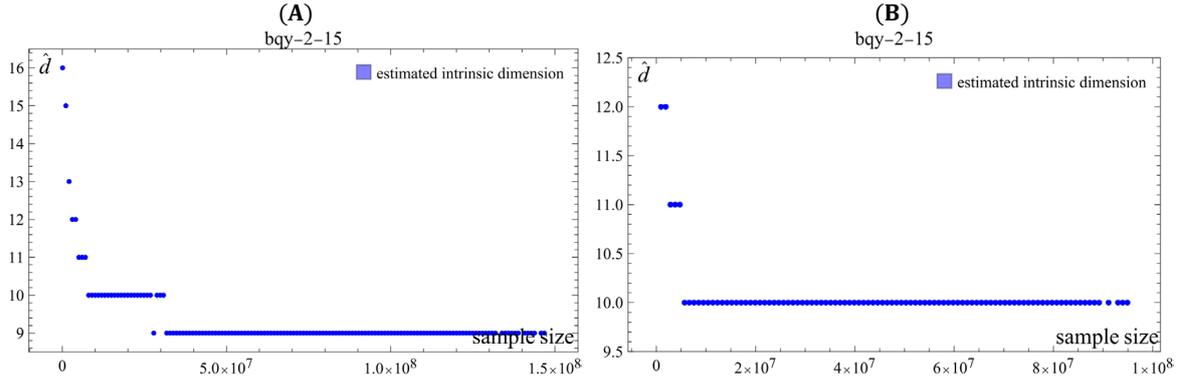

**Fig 5. Dependence of Brito algorithm estimates on sample size for:** (A) SVD vector representations of Russian national literature ($n = 2, d = 15$); (B) SVD vector representations of English national literature ($n = 2, d = 15$).

The boundaries of the upper estimate of the intrinsic dimension are 9 for the Russian language and 10 for the English one, respectively (we use Brito estimate rather for the control). This is in good agreement with the Ref. [23, 32].

## 7   Conclusions and future directions

1. The present paper introduces a novel natural-science object of study – a language fractal structure; we hypothesise that a set of embeddings of all n-grams of a natural language constitutes a representative sample of this fractal set.
2. We ascertain that, for the Russian and English languages, a language structures possess varying non-integer dimensions, and thus are multifractals indeed.
3. Namely, for the English language these dimensions appear to be 8 for words and 10 for bigrams; for the Russian language, 9 for words and 9 for bigrams. Strikingly, for natural languages, the intrinsic dimension turns out be not so large.

### Acknowledgements

The authors are sincerely indebted to Ms. A. Kogan, HSE for natural language texts preprocessing and embedding constructing, for the Russian and English languages.

The authors are indebted to Mr. J. Cumberland, HSE for text-editing and proof-reading.

---

[17] One can attribute slightly results for d=15 to the fact that available samples are not large enough (please compare with results [10]).

This research was supported in part through computational resources of HPC facilities at HSE University.

This article follows from Strategic Project Strategic Project "Human Brain Resilience: Neurocognitive Technologies for Adaptation, Learning, Development and Rehabilitation in a Changing Environment", which is part of Higher School of Economics' development program under the "Priority 2030" academic leadership initiative. The "Priority 2030" initiative is run by Russia's Ministry of Science and Higher Education as part of National Project "Science and Universities".

**Table 1. Intrinsic dimension estimation for languages SVD representations.**

| language | $n$ | $d$ | $\hat{d}_{BQY}$ | min $\hat{d}_{Schw}$ | max $\hat{d}_{Schw}$ | $\alpha$ |
|---|---|---|---|---|---|---|
| RUSSIAN | 1 | 5 | 4-5 | 4.45 | 4.62 | 0-0.72 |
| | | 10 | 5-6 | 5.04 | 7.18 | 0-1.21 |
| | | 15 | 7-8 | 5.98 | 9.57 | 0-1.21 |
| | 2 | 5 | 5 | 3.82 | 6.79 | 0-4.16 |
| | | 10 | 5 | 4.85 | 9.43 | 0-8.28 |
| | | 15 | 9-10 | 5.73 | 10.51 | 0-8.79 |
| ENGLISH | 1 | 5 | 4 | 4.49 | 4.92 | 0-0.6 |
| | | 10 | 6-7 | 5.55 | 7.14 | 0-0.71 |
| | | 15 | 8 | 6.01 | 9.26 | 0-0.71 |
| | 2 | 5 | 5 | 3.24 | 4.51 | 0-2.63 |
| | | 10 | 7 | 3.86 | 5.80 | 0-1.61 |
| | | 15 | 10 | 5.03 | 11.94 | 0-7 |

**Table 2. Intrinsic dimension estimation for languages CBOW representations.**

| language | $n$ | $d$ | $\hat{d}_{BQY}$ | min $\hat{d}_{Schw}$ | max $\hat{d}_{Schw}$ | $\alpha$ |
|---|---|---|---|---|---|---|
| RUSSIAN | 1 | 5 | 5 | 4.65 | 5.52 | 0-4.21 |
| | | 10 | 8 | 7.14 | 8.20 | 0-5.03 |
| | | 15 | 15 | 9.81 | 12.78 | 0-7.33 |
| | 2 | 5 | 5-6 | 5.11 | 7.79 | 0-10 |
| | | 10 | 8-9 | 6.48 | 8.52 | 0-8.88 |
| | | 15 | 13-14 | 8.26 | 10.70 | 0-10 |
| ENGLISH | 1 | 5 | 5 | 4.71 | 5.16 | 0-2.82 |
| | | 10 | 8 | 6.63 | 7.89 | 0-4.02 |
| | | 15 | 14-15 | 9.61 | 12.39 | 0-4.97 |
| | 2 | 5 | 5-6 | 5.23 | 7.82 | 0-5.62 |
| | | 10 | 8-9 | 6.72 | 8.59 | 0-10 |
| | | 15 | 14 | 8.61 | 9.85 | 0-10 |

## Supplementary materials description

Auxiliary information is divided into five sections:

- Appendix A demonstrates the results of testing the Schweinhart and Brito algorithms on synthetic data in terms of absolute percentage error;
- Appendix B contains the results of testing the Schweinhart algorithm for robustness to sample noise;

- Appendix C shows the convergence of the Brito algorithm to a finite value of intrinsic dimension as the sample size increases;
- Appendices D and E display the dependence of intrinsic dimension estimates on vector representations of natural languages, SVD, and CBOW, respectively;

# Supplementary materials

## Appendix A. Absolute percent error (APE) of intrinsic dimension estimation algorithms for conventional geometric objects.

**A1 Table. Absolute percentage error (APE) for a Schweinhart estimator of intrinsic Hausdorff dimension for various geometrical objects.** For each object, the table indicates embedding space dimension $d$ its true intrinsic Hausdorff dimension $d_H$.

| $\alpha$ | Mobius tape APE | Swiss-roll APE | Unit-cube APE | Unit-sphere APE | Unit-sphere (Gauss.) APE | Menger sponge APE | Sierpinski carpet APE |
|---|---|---|---|---|---|---|---|
| 1.00 | 0.47% | 0.43% | 1.35% | 0.07% | 0.68% | 1.56% | 0.07% |
| 2.00 | 0.77% | 0.43% | 1.57% | 0.03% | 1.26% | 1.68% | 0.03% |
| 3.00 | 1.15% | 0.47% | 1.77% | 0.03% | 2.41% | 1.76% | 0.11% |
| 4.00 | 1.59% | 0.55% | 1.98% | 0.08% | 3.79% | 1.82% | 0.18% |
| 5.00 | 2.10% | 0.68% | 2.20% | 0.12% | 5.02% | 1.85% | 0.24% |
| 6.00 | 2.66% | 0.85% | 2.43% | 0.16% | 5.99% | 1.87% | 0.28% |
| 7.00 | 3.22% | 1.08% | 2.69% | 0.19% | 6.77% | 1.90% | 0.32% |
| 8.00 | 3.72% | 1.36% | 2.97% | 0.22% | 7.43% | 1.92% | 0.35% |
| 9.00 | 4.11% | 1.69% | 3.28% | 0.25% | 8.05% | 1.97% | 0.38% |
| 10.00 | 4.33% | 2.04% | 3.62% | 0.29% | 8.58% | 2.03% | 0.42% |
| mean | 2.41% | 0.96% | 2.38% | 0.14% | 5.00% | 1.84% | 0.24% |

**A2 Table. Absolute percentage error (APE) for a Brito estimator of intrinsic Lebesgue dimension for various geometrical objects.** For each object, the table indicates embedding space dimension $d$ its true intrinsic topological dimension $d_T$.

| $\alpha$ | Unit-sphere APE $d=9$ $d_T=2$ | Unit-sphere APE $d=18$ $d_T=5$ | Unit-sphere APE $d=27$ $d_T=8$ | Paraboloid APE $d=9$ $d_T=2$ | Paraboloid APE $d=18$ $d_T=5$ | Paraboloid APE $d=27$ $d_T=8$ | Mobius tape APE $d=3$ $d_T=2$ | Swiss-roll APE $d=3$ $d_T=2$ |
|---|---|---|---|---|---|---|---|---|
| 250 | 0.03% | 17.3% | 32.0% | 0.07% | 8.73% | 1.21% | 0.07% | 49.7% |
| 333 | 0.00% | 16.6% | 13.1% | 0.03% | 54.3% | 24.1% | 0.00% | 0.00% |
| 417 | 0.01% | 5.5% | 18.7% | 0.01% | 2.30% | 28.4% | 0.00% | 0.00% |
| 500 | 1.59% | 12.5% | 15.6% | 0.01% | 17.8% | 40.7% | 0.00% | 0.00% |
| 583 | 0.00% | 17.5% | 11.3% | 0.00% | 23.4% | 4.90% | 0.00% | 0.00% |
| 667 | 0.00% | 1.29% | 19.0% | 0.00% | 36.8% | 36.8% | 0.00% | 0.00% |
| 750 | 0.00% | 4.28% | 8.93% | 0.00% | 30.8% | 39.2% | 0.00% | 0.00% |
| 833 | 0.00% | 15.3% | 24.8% | 0.00% | 30.3% | 32.7% | 0.00% | 0.00% |
| 917 | 0.00% | 9.85% | 8.77% | 0.00% | 4.60% | 28.5% | 0.00% | 0.00% |
| 1000 | 0.00% | 12.1% | 1.45% | 0.00% | 17.0% | 25.6% | 0.00% | 0.00% |
| mean | 0.00% | 11.2% | 15.3% | 0.02% | 22.6% | 26.2% | 0.01% | 4.98% |

# Appendix B. Resistance of the Schweinhart algorithm to noise.

**B1 Table. The influence of noise on the results of the Schweinhart algorithm.** In the case of Gaussian noise, noise was added independently to each coordinate. In the case of uniform noise in a hypercube containing a geometric object, new points were generated from the uniform distribution at a certain percentage $p$ of the original sample size. The biggest impact comes from adding uniform noise points. However, even in this case, the lower bound of the algorithm's estimates corresponds to the intrinsic dimension of the geometric object.

| Noise type | Noise params | $\min \hat{d}_{Schw}$ | $\max \hat{d}_{Schw}$ | $\alpha$ |
|---|---|---|---|---|
| Gaussian (on coordinates) | $\mu = 0 \quad \sigma = 0.001$ | 1.97 | 2.02 | 0-3.4 |
| | $\mu = 0 \quad \sigma = 0.01$ | 1.96 | 2.05 | 0-5 |
| | $\mu = 0 \quad \sigma = 1$ | 3.07 | 3.47 | 0-4.23 |
| Uniform | $p = 0.05$ | 2.07 | 2.91 | 0-2.13 |
| | $p = 0.2$ | 2.01 | 3.04 | 0-5 |

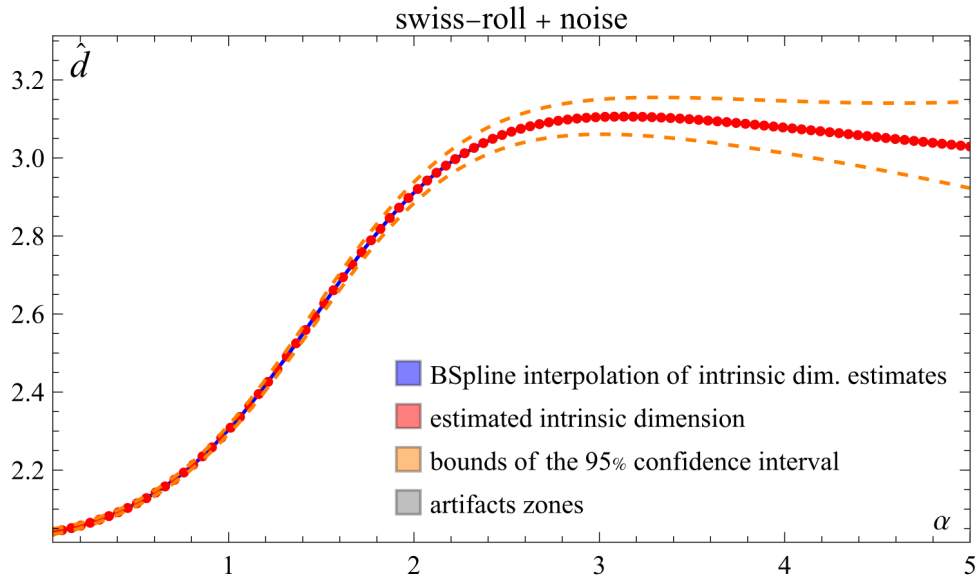

**B2 Fig. Dependence of Schweinhart algorithm estimates on parameter α for the Swiss-roll manifold ($d_H = 2, d = 3$) with a white noise added as a 20% of all points in dataset.** The red dots represent the algorithm's estimates, the blue line is a B-spline interpolation of the dependence of the estimates on α, and the orange dotted line represents the upper and lower limits of the 95% confidence interval. The figure also visualizes the criteria for the informativeness of the parameter α, such as the discrepancy of confidence intervals (grey areas).

**Appendix C. Behavior of the Brito algorithm when reducing the ratio of the intrinsic dimension $d_T$ to the dimension of the embedding space $d$.**

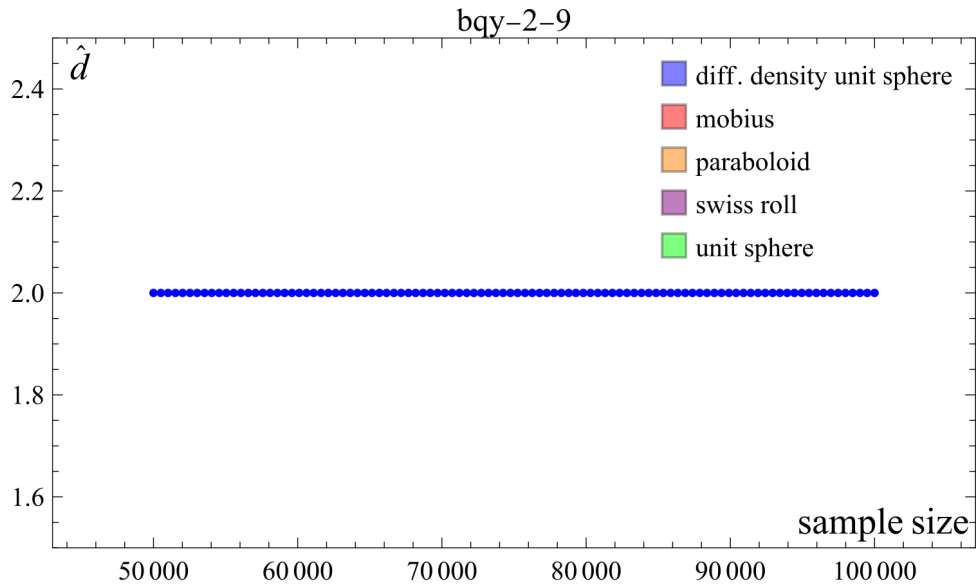

**C1 Fig. Dependence of Brito algorithm estimates on sample size for different manifolds with $d_T = 2$ and $d = 9$.** Algorithm output values are not rounded to show convergence as the sample increases. The algorithm initially converges to a constant intrinsic dimension for all datasets under consideration.

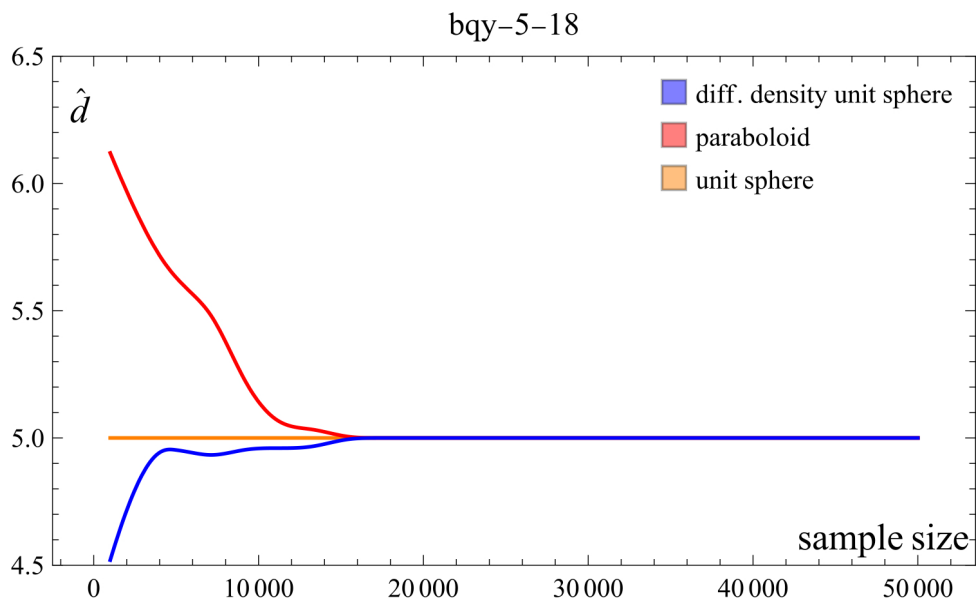

**C2 Fig. Dependence of Brito algorithm estimates on sample size for different manifolds with $d_T = 5$ and $d = 18$.** Algorithm output values are not rounded to show convergence as the sample increases. The algorithm converges to a constant intrinsic dimension for all datasets under consideration.

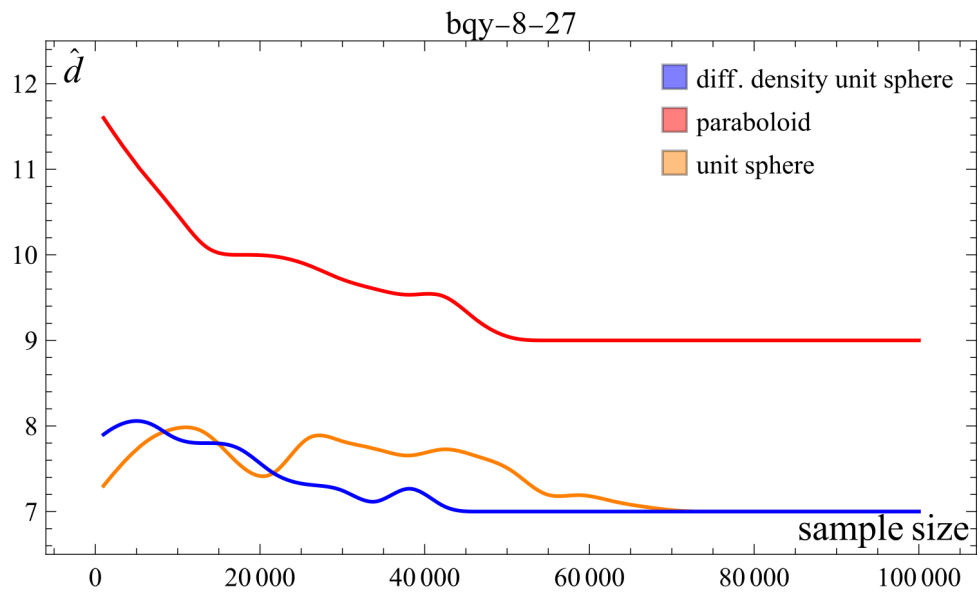

**C3 Fig. Dependence of Brito algorithm estimates on sample size for different manifolds with $d_T = 8$ and $d = 27$.** Algorithm output values are not rounded to show convergence as the sample increases. The algorithm converges to constant intrinsic dimensions for all datasets under consideration.

# Appendix D. Intrinsic dimension estimation for languages SVD representations.

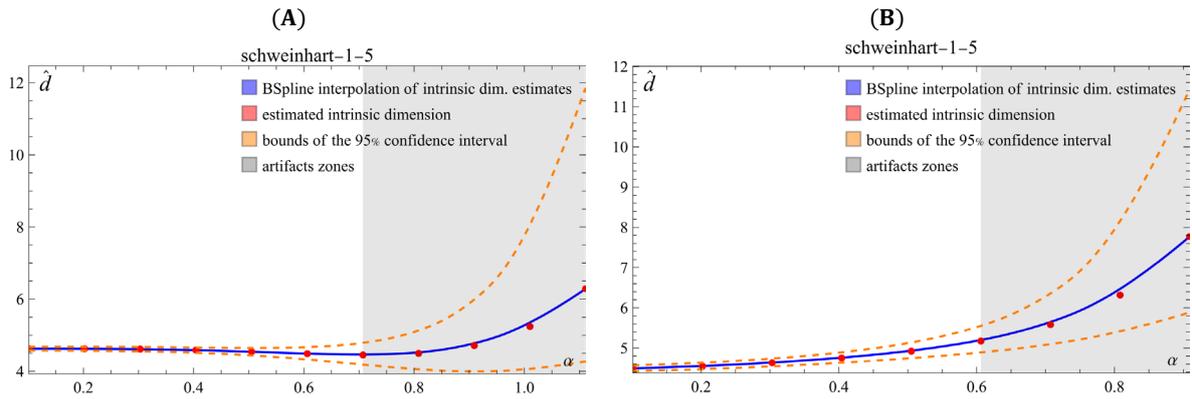

**D1 Fig. Dependence of Schweinhart algorithm estimates on parameter α for:** (A) SVD vector representations of Russian national literature ($n = 1, d = 5$); (B) SVD vector representations of English national literature ($n = 1, d = 5$). The red dots represent the algorithm's estimates, the blue line is a B-spline interpolation of the dependence of the estimates on α, and the orange dotted line represents the upper and lower limits of the 95% confidence interval. The figure also visualizes the criteria for the informativeness of the parameter α, such as the discrepancy of confidence intervals (grey areas).

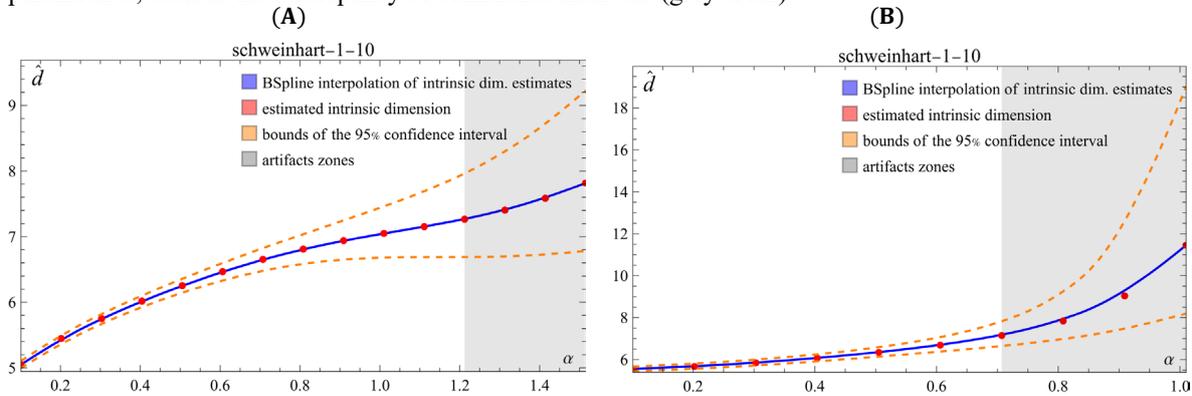

**D2 Fig. Dependence of Schweinhart algorithm estimates on parameter α for:** (A) SVD vector representations of Russian national literature ($n = 1, d = 10$); (B) SVD vector representations of English national literature ($n = 1, d = 10$). The red dots represent the algorithm's estimates, the blue line is a B-spline interpolation of the dependence of the estimates on α, and the orange dotted line represents the upper and lower limits of the 95% confidence interval. The figure also visualizes the criteria for the informativeness of the parameter α, such as the discrepancy of confidence intervals (grey areas).

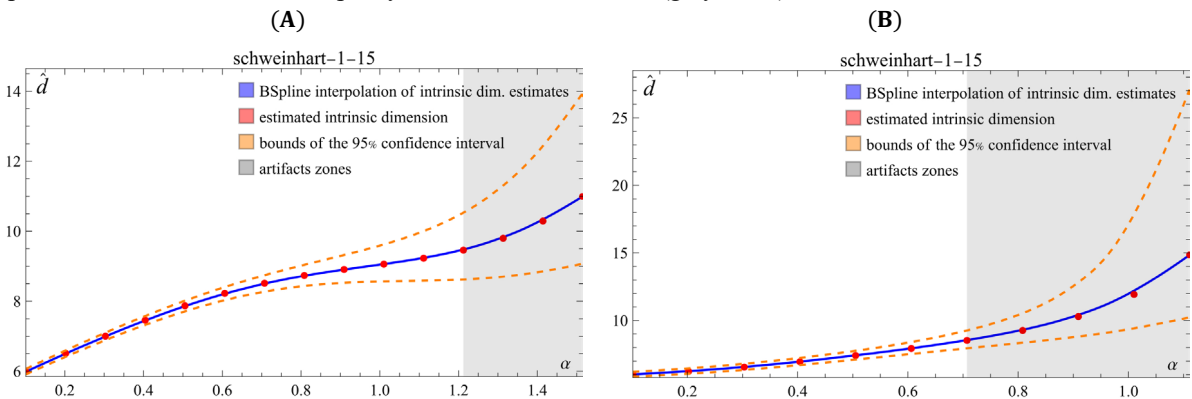

**D3 Fig. Dependence of Schweinhart algorithm estimates on parameter α for:** (A) SVD vector representations of Russian national literature ($n = 1, d = 15$); (B) SVD vector representations of English national literature ($n = 1, d = 15$). The red dots represent the algorithm's estimates, the blue line is a B-spline interpolation of the dependence of the estimates on α, and the orange dotted line represents the upper and lower limits of the 95% confidence interval. The figure also visualizes the criteria for the informativeness of the parameter α, such as the discrepancy of confidence intervals (grey areas).

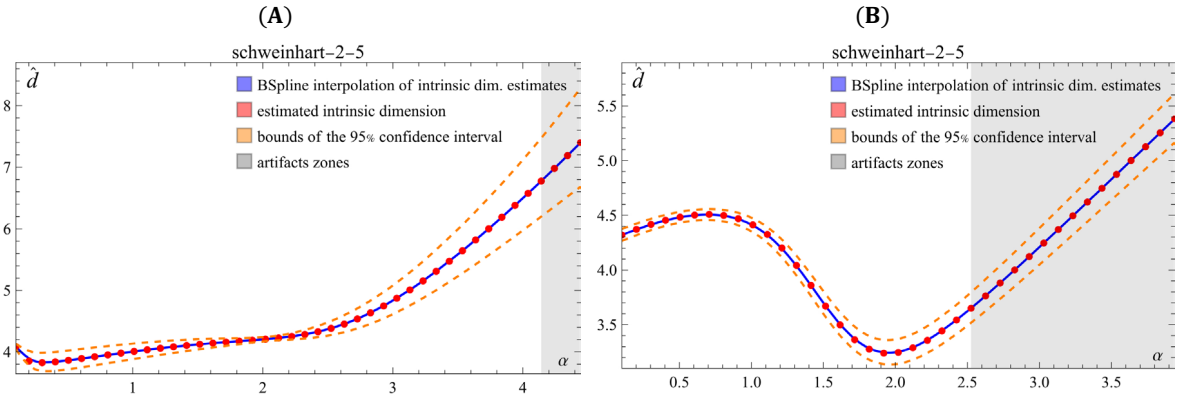

**D4 Fig. Dependence of Schweinhart algorithm estimates on parameter α for:** (A) SVD vector representations of Russian national literature ($n = 2, d = 5$); (B) SVD vector representations of English national literature ($n = 2, d = 5$). The red dots represent the algorithm's estimates, the blue line is a B-spline interpolation of the dependence of the estimates on α, and the orange dotted line represents the upper and lower limits of the 95% confidence interval. The figure also visualizes the criteria for the informativeness of the parameter α, such as the discrepancy of confidence intervals (grey areas).

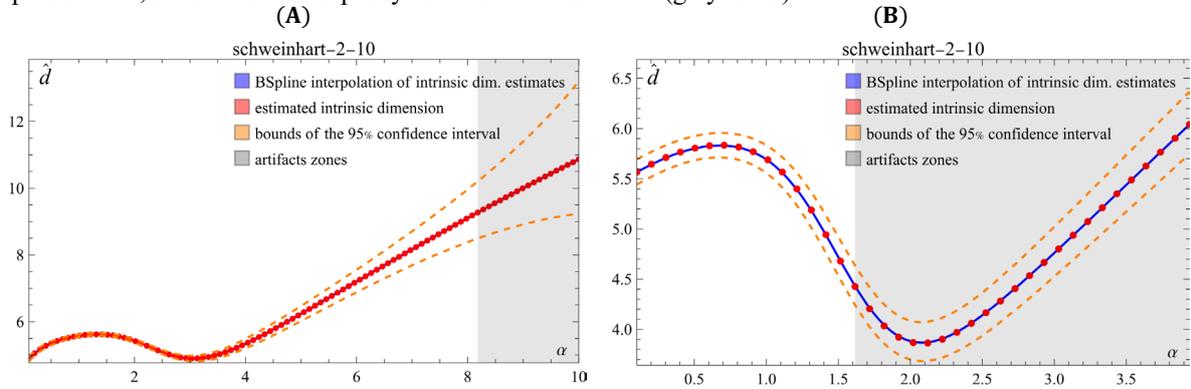

**D5 Fig. Dependence of Schweinhart algorithm estimates on parameter α for:** (A) SVD vector representations of Russian national literature ($n = 2, d = 10$); (B) SVD vector representations of English national literature ($n = 2, d = 10$). The red dots represent the algorithm's estimates, the blue line is a B-spline interpolation of the dependence of the estimates on α, and the orange dotted line represents the upper and lower limits of the 95% confidence interval. The figure also visualizes the criteria for the informativeness of the parameter α, such as the discrepancy of confidence intervals (grey areas).

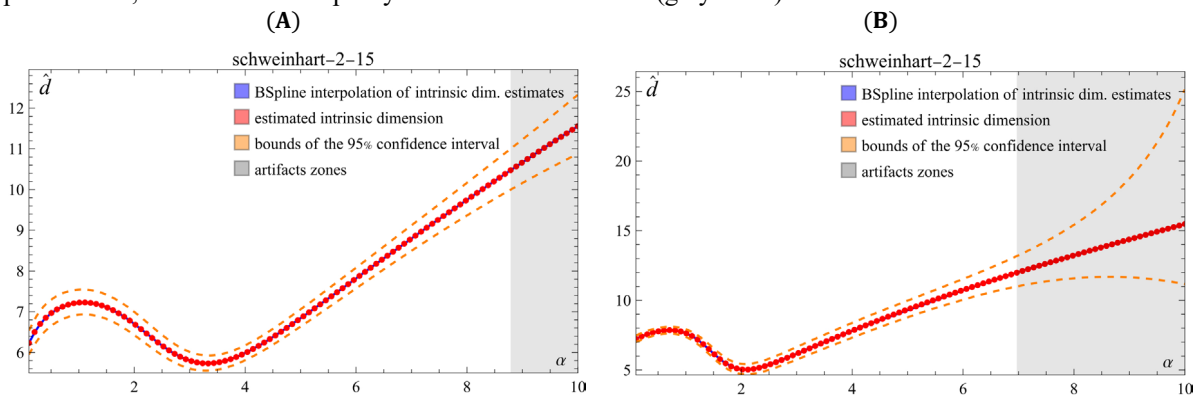

**D6 Fig. Dependence of Schweinhart algorithm estimates on parameter α for:** (A) SVD vector representations of Russian national literature ($n = 2, d = 15$); (B) SVD vector representations of English national literature ($n = 2, d = 15$). The red dots represent the algorithm's estimates, the blue line is a B-spline interpolation of the dependence of the estimates on α, and the orange dotted line represents the upper and lower limits of the 95% confidence interval. The figure also visualizes the criteria for the informativeness of the parameter α, such as the discrepancy of confidence intervals (grey areas).

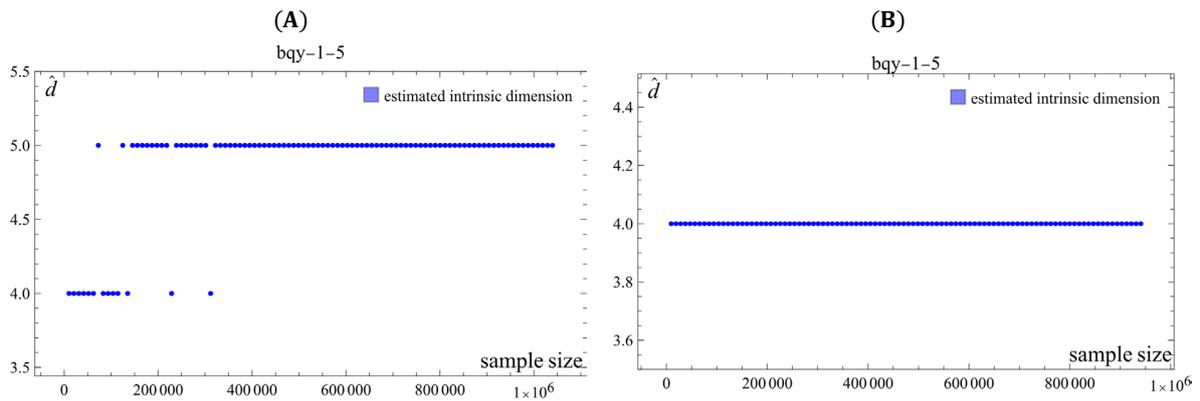

**D7 Fig. Dependence of Brito algorithm estimates on sample size for:** (A) SVD vector representations of Russian national literature ($n = 1, d = 5$); (B) SVD vector representations of English national literature ($n = 1, d = 5$).

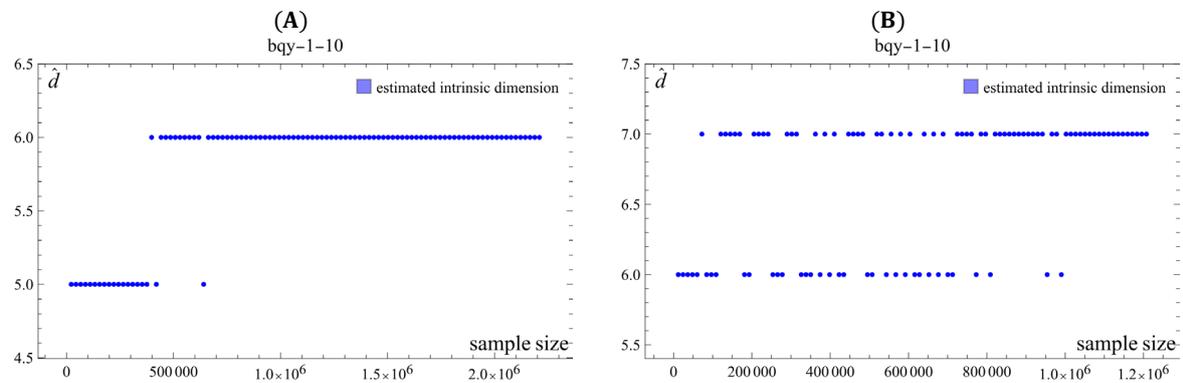

**D8 Fig. Dependence of Brito algorithm estimates on sample size for:** (A) SVD vector representations of Russian national literature ($n = 1, d = 10$); (B) SVD vector representations of English national literature ($n = 1, d = 10$).

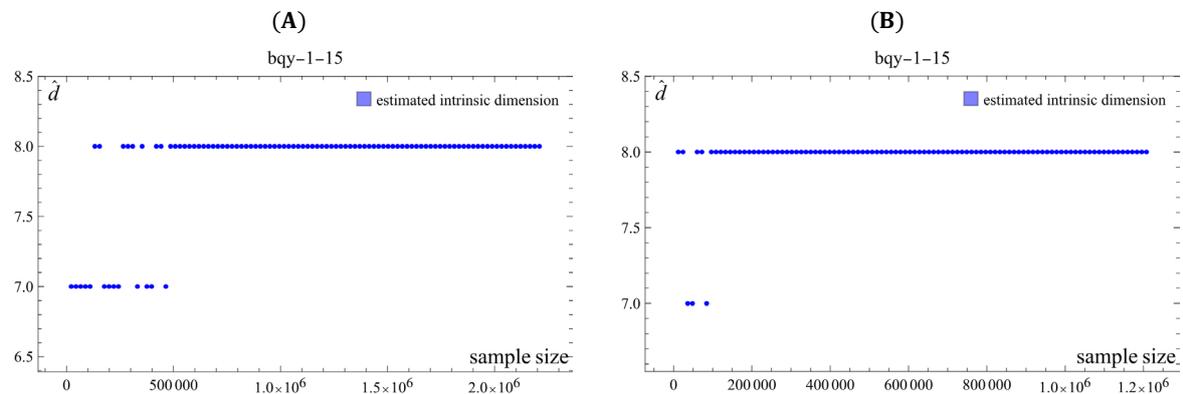

**D9 Fig. Dependence of Brito algorithm estimates on sample size for:** (A) SVD vector representations of Russian national literature ($n = 1, d = 15$); (B) SVD vector representations of English national literature ($n = 1, d = 15$).

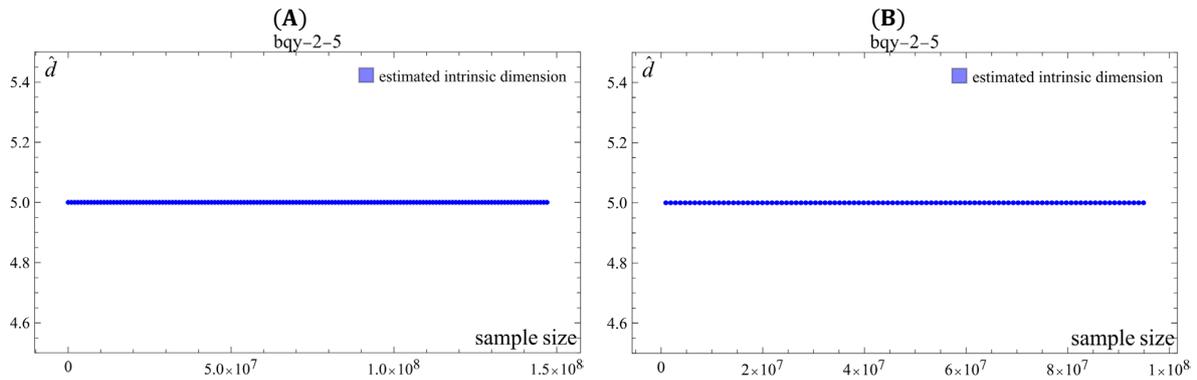

**D10 Fig. Dependence of Brito algorithm estimates on sample size for:** (A) SVD vector representations of Russian national literature ($n = 2, d = 5$); (B) SVD vector representations of English national literature ($n = 2, d = 5$).

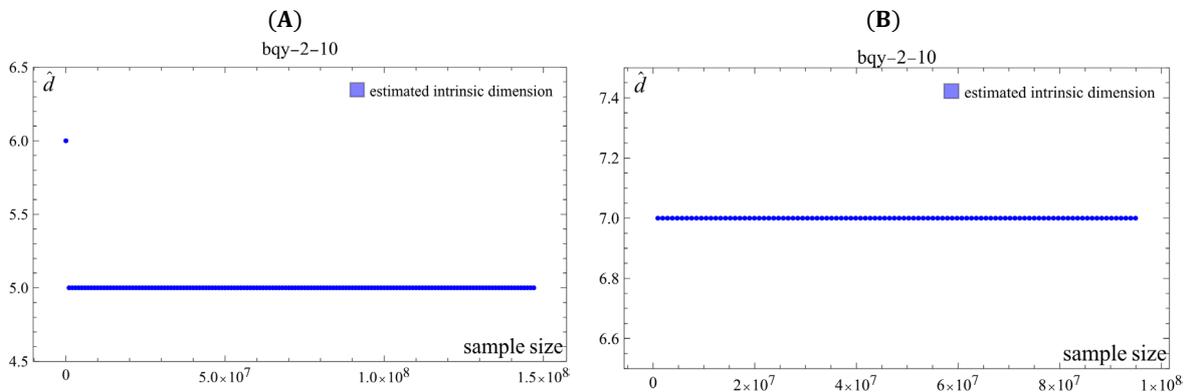

**D11 Fig. Dependence of Brito algorithm estimates on sample size for:** (A) SVD vector representations of Russian national literature ($n = 2, d = 10$); (B) SVD vector representations of English national literature ($n = 2, d = 10$).

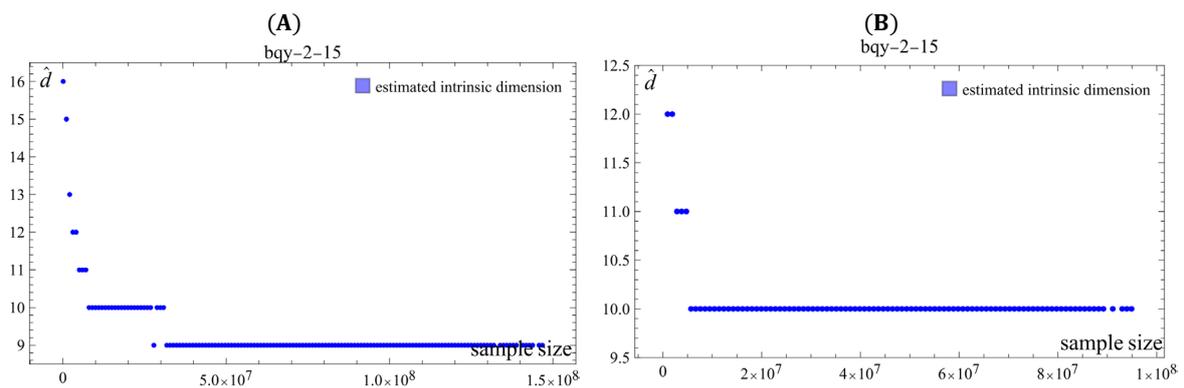

**D12 Fig. Dependence of Brito algorithm estimates on sample size for:** (A) SVD vector representations of Russian national literature ($n = 2, d = 15$); (B) SVD vector representations of English national literature ($n = 2, d = 15$).

# Appendix E. Intrisic dimension estimation for languages CBOW representations.

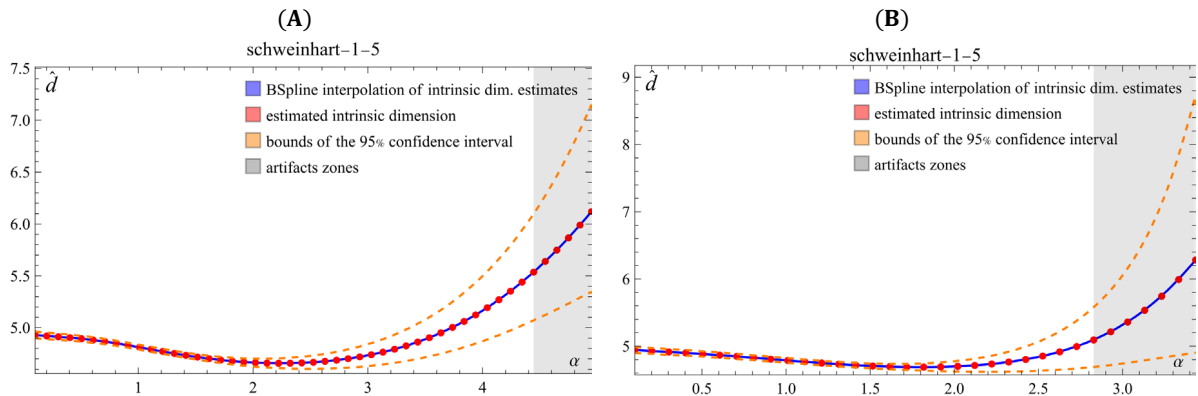

**E1 Fig. Dependence of Schweinhart algorithm estimates on parameter α for:** (A) CBOW vector representations of Russian national literature ($n = 1, d = 5$); (B) CBOW vector representations of English national literature ($n = 1, d = 5$). The red dots represent the algorithm's estimates, the blue line is a B-spline interpolation of the dependence of the estimates on α, and the orange dotted line represents the upper and lower limits of the 95% confidence interval. The figure also visualizes the criteria for the informativeness of the parameter α, such as the discrepancy of confidence intervals (grey areas).

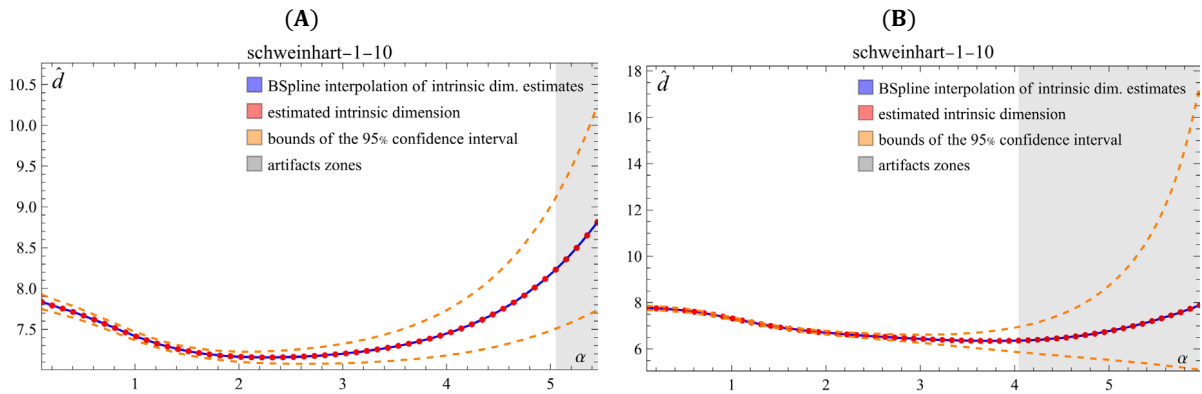

**E2 Fig. Dependence of Schweinhart algorithm estimates on parameter α for:** (A) CBOW vector representations of Russian national literature ($n = 1, d = 10$); (B) CBOW vector representations of English national literature ($n = 1, d = 10$). The red dots represent the algorithm's estimates, the blue line is a B-spline interpolation of the dependence of the estimates on α, and the orange dotted line represents the upper and lower limits of the 95% confidence interval. The figure also visualizes the criteria for the informativeness of the parameter α, such as the discrepancy of confidence intervals (grey areas).

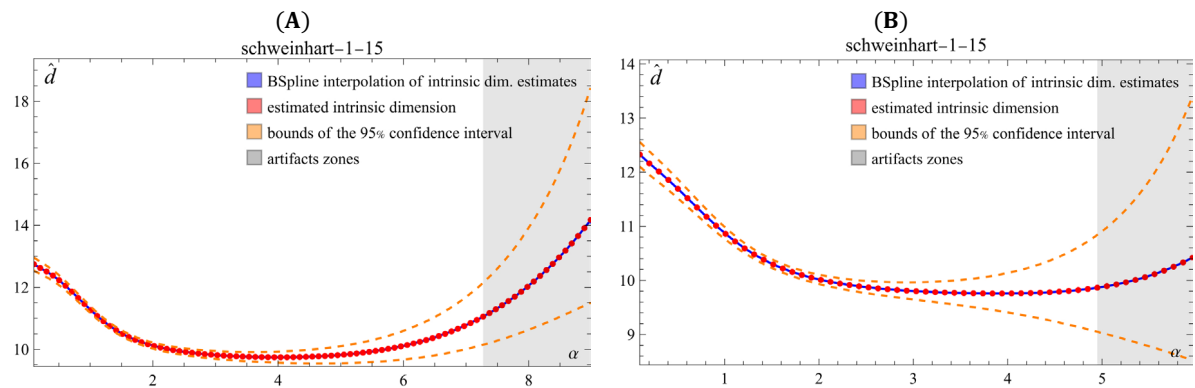

**E3 Fig. Dependence of Schweinhart algorithm estimates on parameter α for:** (A) CBOW vector representations of Russian national literature ($n = 1, d = 15$); (B) CBOW vector representations of English national literature ($n = 1, d = 15$). The red dots represent the algorithm's estimates, the blue line is a B-spline interpolation of the dependence of the estimates on α, and the orange dotted line represents the upper and lower limits of the 95% confidence interval. The figure also visualizes the criteria for the informativeness of the parameter α, such as the discrepancy of confidence intervals (grey areas).

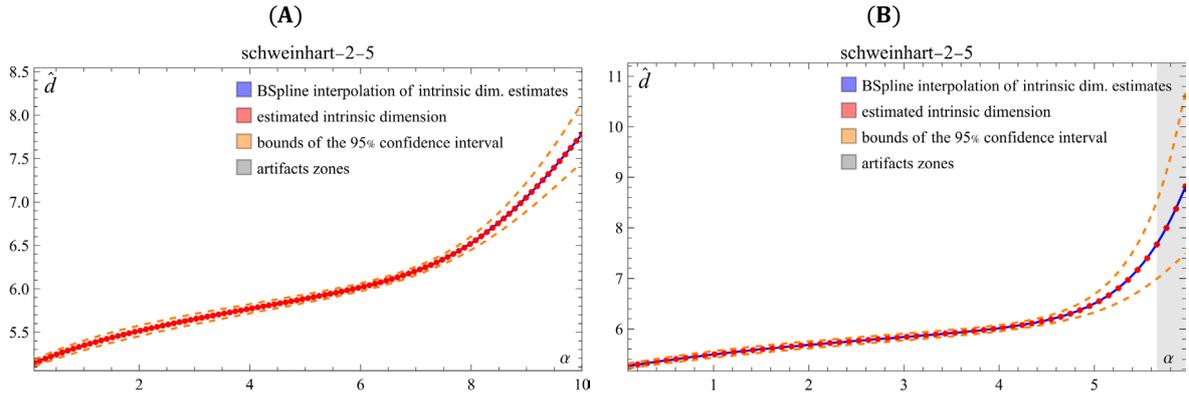

**E4 Fig. Dependence of Schweinhart algorithm estimates on parameter α for:** (A) CBOW vector representations of Russian national literature ($n = 2, d = 5$); (B) CBOW vector representations of English national literature ($n = 2, d = 5$). The red dots represent the algorithm's estimates, the blue line is a B-spline interpolation of the dependence of the estimates on α, and the orange dotted line represents the upper and lower limits of the 95% confidence interval. The figure also visualizes the criteria for the informativeness of the parameter α, such as the discrepancy of confidence intervals (grey areas).

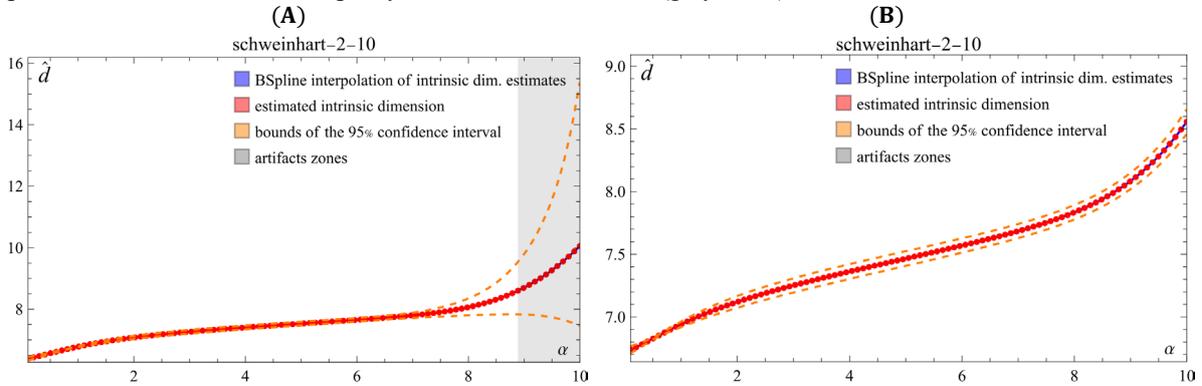

**E5 Fig. Dependence of Schweinhart algorithm estimates on parameter α for:** (A) CBOW vector representations of Russian national literature ($n = 2, d = 10$); (B) CBOW vector representations of English national literature ($n = 2, d = 10$). The red dots represent the algorithm's estimates, the blue line is a B-spline interpolation of the dependence of the estimates on α, and the orange dotted line represents the upper and lower limits of the 95% confidence interval. The figure also visualizes the criteria for the informativeness of the parameter α, such as the discrepancy of confidence intervals (grey areas).

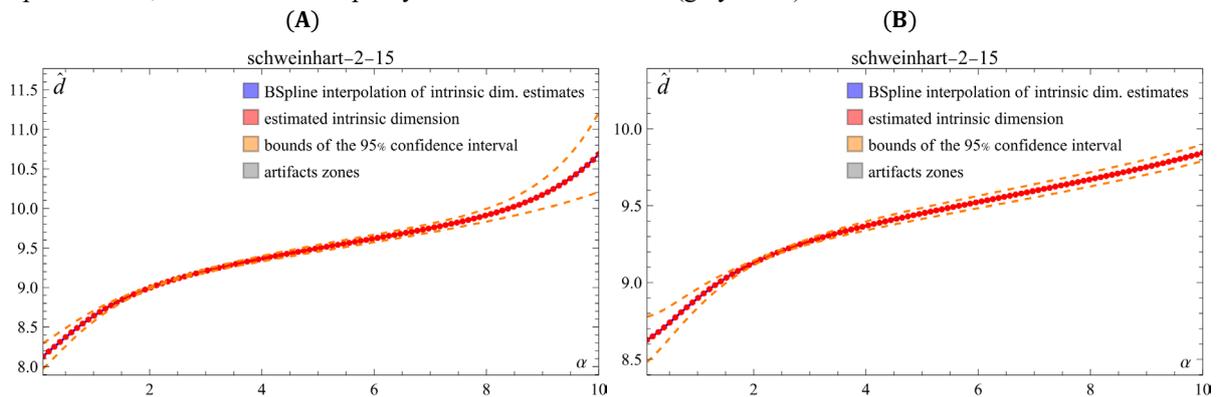

**E6 Fig. Dependence of Schweinhart algorithm estimates on parameter α for:** (A) CBOW vector representations of Russian national literature ($n = 2, d = 15$); (B) CBOW vector representations of English national literature ($n = 2, d = 15$). The red dots represent the algorithm's estimates, the blue line is a B-spline interpolation of the dependence of the estimates on α, and the orange dotted line represents the upper and lower limits of the 95% confidence interval. The figure also visualizes the criteria for the informativeness of the parameter α, such as the discrepancy of confidence intervals (grey areas).

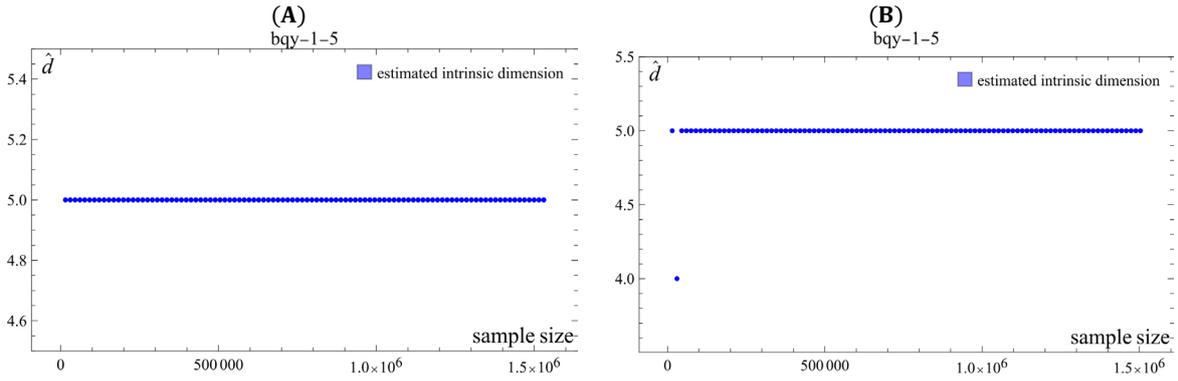

**E7 Fig. Dependence of Brito algorithm estimates on sample size for:** (A) CBOW vector representations of Russian national literature ($n = 1, d = 5$); (B) CBOW vector representations of English national literature ($n = 1, d = 5$).

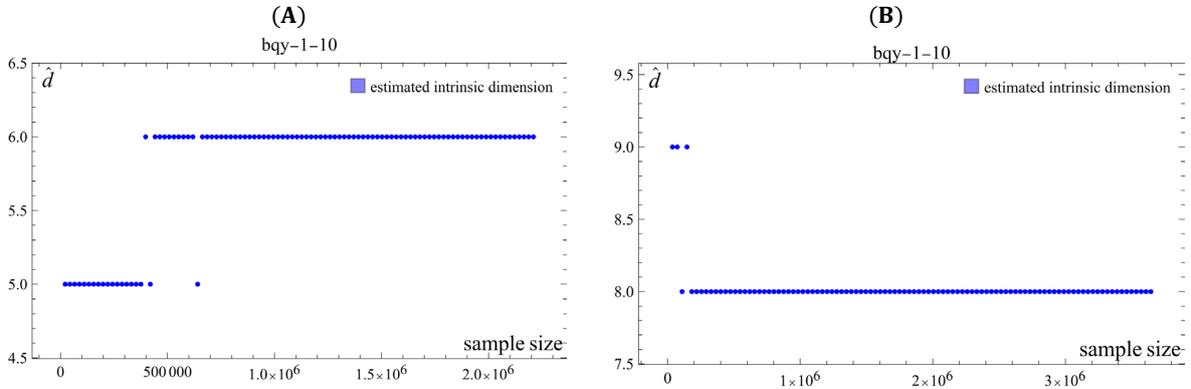

**E8 Fig. Dependence of Brito algorithm estimates on sample size for:** (A) CBOW vector representations of Russian national literature ($n = 1, d = 10$); (B) CBOW vector representations of English national literature ($n = 1, d = 10$).

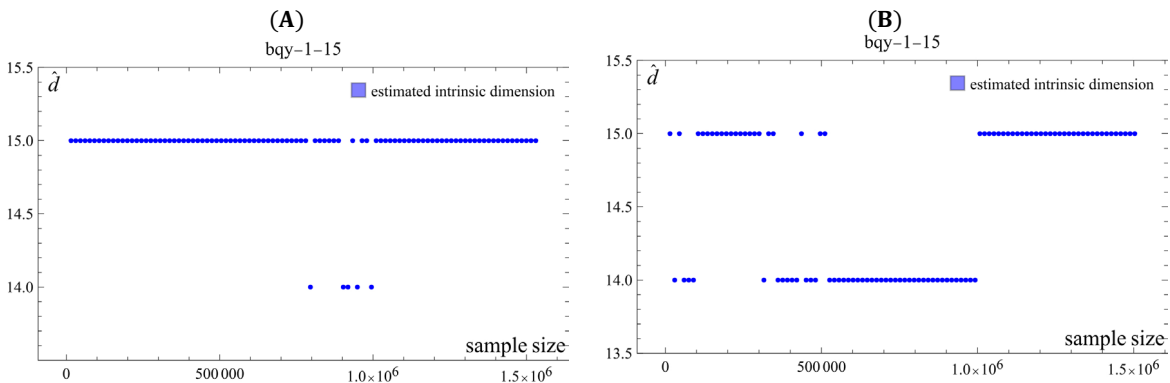

**E9 Fig. Dependence of Brito algorithm estimates on sample size for:** (A) CBOW vector representations of Russian national literature ($n = 1, d = 15$); (B) CBOW vector representations of English national literature ($n = 1, d = 15$).

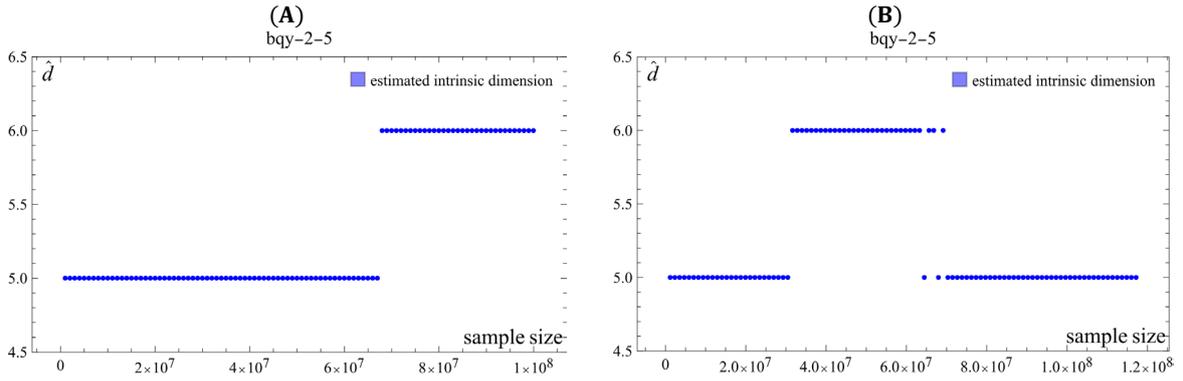

**E10 Fig. Dependence of Brito algorithm estimates on sample size for:** (A) CBOW vector representations of Russian national literature ($n = 2, d = 5$); (B) CBOW vector representations of English national literature ($n = 2, d = 5$).

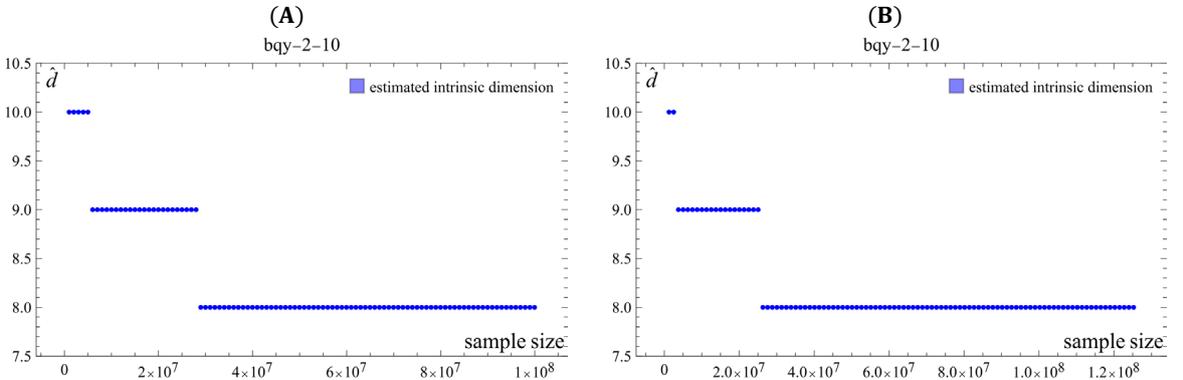

**E11 Fig. Dependence of Brito algorithm estimates on sample size for:** (A) CBOW vector representations of Russian national literature ($n = 2, d = 10$); (B) CBOW vector representations of English national literature ($n = 2, d = 10$).

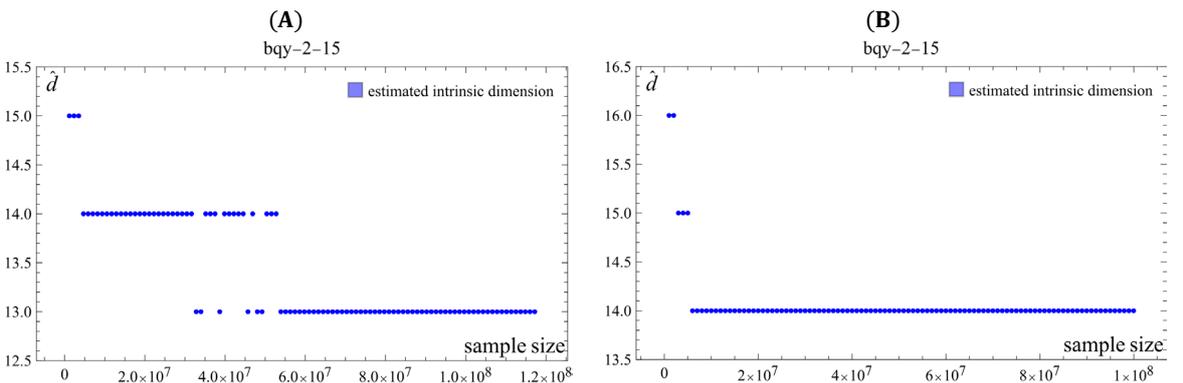

**E12 Fig. Dependence of Brito algorithm estimates on sample size for:** (A) CBOW vector representations of Russian national literature ($n = 2, d = 15$); (B) CBOW vector representations of English national literature ($n = 2, d = 15$).